# X2Face: A network for controlling face generation using images, audio, and pose codes

Olivia Wiles*, A. Sophia Koepke*, Andrew Zisserman

Visual Geometry Group,
University of Oxford
{ow,koepke,az}@robots.ox.ac.uk

**Abstract.** The objective of this paper is a neural network model that controls the pose and expression of a given face, using another face or modality (e.g. audio). This model can then be used for lightweight, sophisticated video and image editing.

We make the following three contributions. First, we introduce a network, X2Face, that can control a *source* face (specified by one or more frames) using another face in a *driving* frame to produce a *generated* frame with the identity of the *source* frame but the pose and expression of the face in the *driving* frame. Second, we propose a method for training the network fully self-supervised using a large collection of video data. Third, we show that the generation process can be driven by other modalities, such as audio or pose codes, without any further training of the network.

The generation results for driving a face with another face are compared to state-of-the-art self-supervised/supervised methods. We show that our approach is more robust than other methods, as it makes fewer assumptions about the input data. We also show examples of using our framework for video face editing.

## 1 Introduction

Being able to animate a still image of a face in a controllable, lightweight manner has many applications in image editing/enhancement and interactive systems (e.g. animating an on-screen agent with natural human poses/expressions). This is a challenging task, as it requires representing the face (e.g. modelling in 3D) in order to control it and a method of mapping the desired form of control (e.g. expression or pose) back onto the face representation. In this paper we investigate whether it is possible to forgo an explicit face representation and instead implicitly learn this in a self-supervised manner from a large collection of video data. Further, we investigate whether this implicit representation can then be used directly to control a face with another modality, such as audio or pose information.

To this end, we introduce X2Face, a novel self-supervised network architecture that can be used for face puppeteering of a *source* face given a *driving vector*.

---

* Denotes equal contribution.

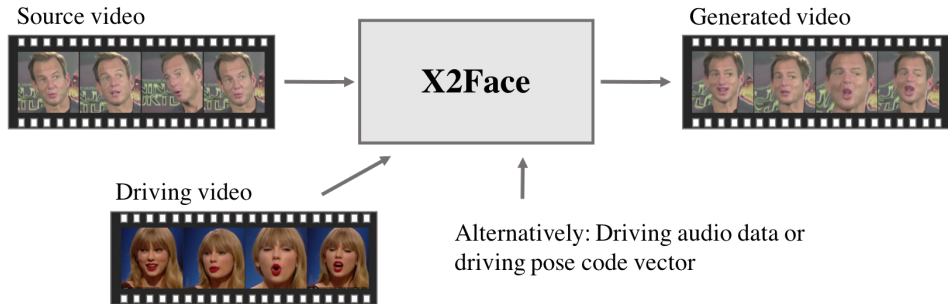

Fig. 1: Overview of X2Face: a model for controlling a *source* face using a *driving* frame, audio data, or specifying a pose vector. X2Face is trained without expression or pose labels.

The *source* face is instantiated from a single or multiple *source* frames, which are extracted from the same face track. The *driving vector* may come from multiple modalities: a *driving* frame from the same or another video face track, pose information, or audio information; this is illustrated in Fig. 1. The *generated* frame resulting from X2Face has the identity, hairstyle, etc. of the *source* face but the properties of the *driving vector* (e.g. the given pose, if pose information is given; or the *driving* frame's expression/pose, if a *driving* frame is given).

The network is trained in a self-supervised manner using pairs of *source* and *driving* frames. These frames are input to two subnetworks: the *embedding network* and the *driving network* (see Fig. 2). By controlling the information flow in the network architecture, the model learns to factorise the problem. The *embedding network* learns an *embedded* face representation for the *source* face – effectively face frontalisation; the *driving network* learns how to map from this *embedded* face representation to the *generated* frame via an embedding, named the *driving vector*.

The X2Face network architecture is described in Section 3.1, and the self-supervised training framework in Section 3.2. In addition we make two further contributions. First, we propose a method for linearly regressing from a set of labels (e.g. for head pose) or features (e.g. from audio) to the *driving vector*; this is described in Section 4. The performance is evaluated in Section 5, where we show (i) the robustness of the generated results compared to state-of-the-art self-supervised [45] and supervised [1] methods; and (ii) the controllability of the network using other modalities, such as audio or pose. The second contribution, described in Section 6, shows how the *embedded* face representation can be used for video face editing, e.g. adding facial decorations in the manner of [31] using multiple or just a single *source* frame.

## 2   Related work

**Explicit modelling of faces for image generation.** Traditionally facial animation (or puppeteering) given one image was performed by fitting a 3DMM and then modifying the estimated parameters [3]. Later work has built on the fitting of 3DMMs by including high level details [34,41], taking into account additional images [33] or 3D scans [4], or by learning 3DMM parameters directly from RGB data without ground truth labels [39,2]. Please refer to Zollhöfer et. al. [46] for a survey.

Given a driving and source video sequence, a 3DMM or 3D mesh can be obtained and used to model both the driving and source face [43,40,10]. The estimated 3D is used to transform the expression of the source face to match that of the driving face. However, this requires additional steps to transfer the hidden regions (e.g. the teeth). As a result, a neural network conditioned on a single driving image can be used to predict higher level details to fill in these hidden regions [25].

Motivated by the fact that a 3DMM approach is limited by the components of the corresponding morphable model, which may not model the full range of required expressions/deformations and the higher level details, [1] propose a 2D warping method. Given only one source image, [1] use facial landmarks in order to warp the expression of one face onto another. They additionally allow for fine scale details to be transferred by monitoring changes in the driving video.

An interesting related set of works consider how to frontalise a face in a still image using a generic reference face [14], transferring expressions of an actor to an avatar [35] and swapping one face with another [20,24].

**Learning based approaches for image generation.** There is a wealth of literature on supervised/self-supervised approaches; here we review only the most relevant work. Supervised approaches for controlling a given face learn to model factors of variation (e.g. lighting, pose, etc.) by conditioning the generated image on known ground truth information which may be head pose, expression, or landmarks [44,21,42,5,12,30]. This requires a training dataset with known pose or expression information which may be expensive to obtain or require subjective judgement (e.g. in determining the expression). Consequently, self-supervised and unsupervised approaches attempt to automatically learn the required factors of variation (e.g. optical flow or pose) without labelling. This can be done by maximising mutual information [7] or by training the network to synthesise future video frames [29,11].

Another relevant self-supervised method is CycleGAN [45] which learns to transform images of one domain into those of another. While not explicitly devised for this task, as CycleGAN learns to be cycle-consistent, the transformed images often bear semantic similarities to the original images. For example, a CycleGAN model trained to transform images of one person's face (domain A) into those of another (domain B), will often learn to map the pose/position/expression of the face in domain A onto the generated face from domain B.

**Using multi-modal setups to control image generation.** Other modalities, such as audio, can control image generation by using a neural network that learns the relationship between audio and correlated parts in corresponding images. Examples are controlling the mouth with speech [8,38], controlling a head with audio and a known emotional state [16], and controlling body movement with music [36].

Our method has the benefits of being self-supervised and the ability to control the generation process from other modalities without requiring explicit modelling of the face. Thus it is applicable to other domains.

## 3   Method

This section introduces the network architecture in Section 3.1, followed by the curriculum strategy used to train the network in Section 3.2.

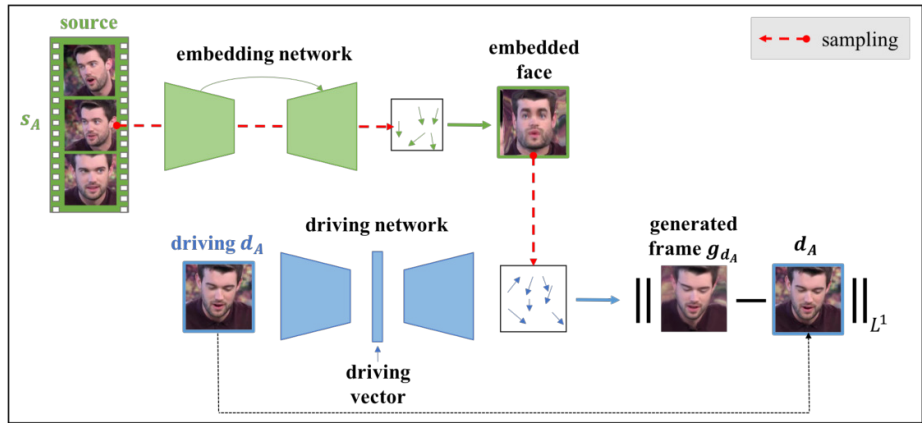

Fig. 2: An overview of X2Face during the initial training stage. Given multiple frames of a video (here 4 frames), one frame is designated the *source* frame and another the *driving* frame. The *source* frame is input to the *embedding network*, which learns a sampler to map pixels from the *source* frame to the *embedded* face. The *driving* frame is input to the *driving network*, which learns to map pixels from the *embedded* face to the *generated* frame. The *generated* frame should have the identity of the *source* frame and the pose/expression of the *driving* frame. In this training stage, as the frames are from the same video, the *generated* and *driving* frames should match. However, at test time the identities of the *source* and *driving* face can differ.

### 3.1   Architecture

The network takes two inputs: a *driving* and a *source* frame. The *source* frame is input to the *embedding network* and the *driving* frame to the *driving network*. This is illustrated in Fig. 2. Precise architectural details are given in the supplementary material in Section A.1 and Section A.2.

**Embedding network.** The *embedding network* learns a bilinear sampler to determine how to map from the *source* frame to a face representation, the *embedded* face. The architecture is based on U-Net [32] and pix2pix [15]; the output is a 2-channel image (of the same dimensions as the *source* frame) that encodes the flow $\delta x, \delta y$ for each pixel.

While the *embedding network* is not explicitly forced to frontalise the *source* frame, we observe that it learns to do so for the following reason. Because the *driving network* samples from the *embedded* face to produce the *generated* frame without knowing the pose/expression of the *source* frame, it needs the *embedded* face to have a common representation (e.g. be frontalised) across *source* frames with differing poses and expressions.

**Driving network.** The *driving network* takes a *driving* frame as input and learns a bilinear sampler to transform pixels from the *embedded* face to produce the *generated* frame. It has an encoder-decoder architecture. In order to sample correctly from the *embedded* face and produce the *generated* frame, the latent embedding (the *driving vector*) must encode pose/expression/zoom/other factors of variation.

### 3.2   Training the network

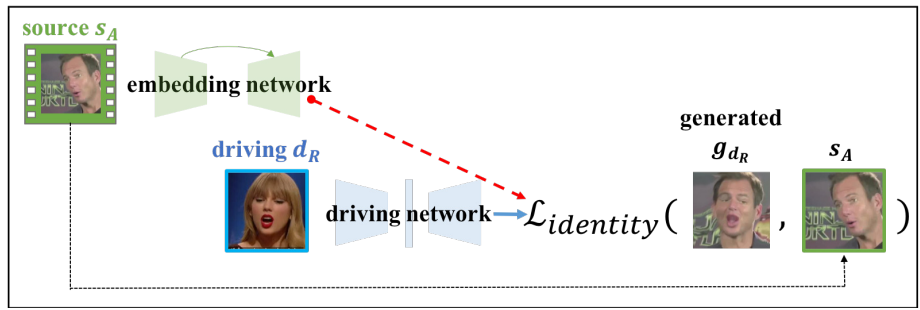

Fig. 3: The identity loss function when the *source* and *driving* frames are of different identities. This loss enforces that the *generated* frame has the same identity as the *source* frame.

The network is trained with a curriculum strategy using two stages. The first training stage (**I**) is fully self-supervised. In the second training stage (**II**), we make use of a CNN pre-trained for face identification to add additional constraints based on the identity of the faces in the *source* and *driving* frames to finetune the model following training stage (**I**).

**I**. The first stage (illustrated in Fig. 2) uses only a pixelwise $L1$ loss between the *generated* and the *driving* frames. Whilst this is sufficient to train the network such that the *driving* frame encodes expression and pose, we observe that some face shape information is leaked through the *driving vector* (e.g. the *generated* face becomes fatter/longer depending on the face in the *driving* frame). Consequently, we introduce additional loss functions – called identity loss functions – in the second stage.

**II**. In the second stage, the identity loss functions are applied to enforce that the identity is the same between the *generated* and the *source* frames irrespective of the identity of the *driving* frame. This loss should mitigate against the face shape leakage discussed in stage **I**. In practice, one *source* frame $s_A$ of identity

A, and two *driving* frames $d_A$,$d_R$ are used as training inputs; $d_A$ is of identity A and $d_R$ a random identity. This gives two generated frames $g_{d_A}$, $g_{d_R}$ respectively, which should both be of identity A. Two identity loss functions are then imposed: $\mathcal{L}_{\text{identity}}(d_A, g_{d_A})$ and $\mathcal{L}_{\text{identity}}(s_A, g_{d_R})$. $\mathcal{L}_{\text{identity}}$ is implemented using a network pre-trained for identity to measure the similarity of the images in feature space by comparing appropriate layers of the network (i.e. a content loss as in [13,6]). The precise layers are chosen based on whether we are considering $g_{d_A}$ or $g_{d_R}$:

1. $\mathcal{L}_{identity}(d_A, g_{d_A})$. $g_{d_A}$ should have the same identity, pose and expression as $d_A$ so we use the photometric $L1$ loss and a $L1$ content loss on the Conv2-5 and Conv7 layers (i.e. layers that encode both lower/higher level information such as pose/identity) between $g_{d_A}$ and $d_A$.

2. $\mathcal{L}_{identity}(s_A, g_{d_R})$ *(Fig. 3)*. $g_{d_R}$ should have the identity of $s_A$ but the pose and expression of $d_R$. Consequently, we cannot use the photometric loss but only a content loss. We minimise a $L1$ content loss on the Conv6-7 layers (i.e. layers encoding higher level identity information) between $g_{d_A}$ and $s_A$.

The pre-trained network used for these losses is the 11-layer VGG network (configuration A) [37] trained on the VGG-Face Dataset [26].

## 4 Controlling the image generation with other modalities

Given a trained X2Face network, the *driving vector* can be used to control the *source* face with other modalities such as audio or pose.

### 4.1 Pose

Instead of controlling the generation with a *driving* frame, we can control the head pose of the *source* face using a pose code such that when varying the code's pitch/yaw/roll angles, the *generated* frame varies accordingly. This is done by learning a forward mapping $f_{p \to v}$ from head pose $p$ to the *driving vector* $v$ such that $f_{p \to v}(p)$ can serve as a modified input to the *driving network*'s decoder. However, this is an ill-posed problem; directly using this mapping loses information, as the *driving vector* encodes more than just pose.

As a result, we use vector arithmetic. Effectively we drive a *source* frame with itself but modify the corresponding *driving vector* $v_{emb}^{source}$ to remove the pose of the *source* frame $p_{source}$ and incorporate the new driving pose $p_{driving}$. This gives:

$$v_{emb}^{driving} = v_{emb}^{source} + v_{emb}^{\Delta pose} = v_{emb}^{source} + f_{p \to v}(p_{driving} - p_{source}). \qquad (1)$$

However, VoxCeleb [23] does not contain ground truth head pose, so an additional mapping $f_{v \to p}$ is needed to determine $p_{source} = f_{v \to p}(v_{emb}^{source})$.

$f_{v \to p}$. $f_{v \to p}$ is trained to regress $p$ from $v$. It is implemented using a fully connected layer with bias and trained using an L1 loss. Training pairs $(v, p)$ are obtained using an annotated dataset with image to pose labels $p$; $v$ is obtained by passing the image through the encoder of the *driving network*.

$f_{p \to v}$. $f_{p \to v}$ is trained to regress $v$ from $p$. It is implemented using a fully-connected linear layer with bias followed by batch-norm. When $f_{v \to p}$ is known, this function can be learnt directly on VoxCeleb by passing an image through X2Face to get the *driving vector* $v$ and $f_{v \to p}(v)$ gives the pose $p$.

### 4.2  Audio

Audio data from the videos in the VoxCeleb dataset can be used to drive a *source* face in a manner similar to that of pose by driving the *source* frame with itself but modifying the *driving vector* using the audio from another frame. The forward mapping $f_{a \to v}$ from audio features $a$ to the corresponding *driving vector* $v$ is trained using pairs of audio features $a$ and driving vectors $v$. These can be directly extracted from VoxCeleb (so no backward mapping $f_{v \to a}$ is required). $a$ is obtained by extracting the 256D audio features from the neural network in [9] and the 128D $v$ by passing the corresponding frame through the *driving network*'s encoder. Ordinary least squares linear regression is then used to learn $f_{a \to v}$ after first normalising the audio features to $\sim N(0, 1)$. No normalisation is used when employing the mapping to drive the frame generation; this amplifies the signal, visually improving the generated results.

As learning the function $f_{a \to v} : \mathbb{R}^{1 \times 256} \to \mathbb{R}^{1 \times 128}$ is under-constrained, the embedding learns to encode some pose information. Therefore, we additionally use the mappings $f_{p \to v}$ and $f_{v \to p}$ described in Section 4.1 to remove this information. Given driving audio features $a_{driving}$ and the corresponding, non-modified *driving vector* $v_{emb}^{source}$, the new *driving vector* $v_{emb}^{driving}$ is then

$$v_{emb}^{driving} = v_{emb}^{source} + f_{a \to v}(a_{driving}) - f_{a \to v}(a_{source}) + f_{p \to v}(p_{audio} - p_{source}),$$

where $p_{source} = f_{v \to p}(v_{emb}^{source})$ is the head pose of the frame input to the *driving network* (i.e. the *source* frame), $p_{audio} = f_{v \to p}(f_{a \to v}(a_{driving}))$ is the pose information contained in $f_{a \to v}(a_{driving})$, and $a_{source}$ is the audio feature vector corresponding to the *source* frame.

## 5  Experiments

This section evaluates X2Face by first performing an ablation study in Section 5.1 on the architecture and losses used for training, followed by results for controlling a face with a *driving* frame in Section 5.2, pose information in Section 5.3, and audio information in Section 5.4.

**Training.** X2Face is trained on the VoxCeleb video dataset [23] using dlib [18] to crop the faces to $256 \times 256$. The identities are randomly split into train/val/test identities (with a split of 75/15/10) and frames extracted at one fps to give 900,764 frames for training and 125,131 frames for testing.

The model is trained in PyTorch [27] using SGD with momentum 0.9 and batch-size of 16. First, it is trained just with $L1$ loss, and a learning rate of 0.001. The learning rate is decreased by a factor of 10 when the loss plateaus. Once the loss converges, the identity losses are incorporated and are weighted as follows: (i) for same identities to be as strong as the photometric $L1$ loss at each layer; (ii) for different identities to be 1/10 the size of the photometric loss at each layer. This training phase is started with a learning rate of 0.0001.

**Testing.** The model can be tested using either a single or multiple *source* frames. The reasoning for this is that if the *embedded* face is stable (e.g. different facial regions always map to the same place on the *embedded* face), we expect to be able to combine multiple *source* frames by averaging over the *embedded* faces.

### 5.1    Architecture studies

To quantify the utility of using additional views at *test* time and the benefit of the curriculum strategy for training the network (i.e. using the identity losses explained in Section 3.2), we evaluate the results for these different settings on a left-out test set of VoxCeleb. We consider 120K *source* and *driving* pairs where the *driving* frame is from the same video as the *source* frames; thus, the *generated* frame should be the same as the *driving* frame. The results are given in Table 1.

Table 1: $L1$ reconstruction error on the test set, comparing the *generated* frame to the ground truth frame (in this case the *driving* frame) for different training/testing setups. Lower is better for $L1$ error. Additionally, we give the percentage improvement over the $L1$ error for the model trained with only training stage **I** and tested with a single *source* frame. In this case, higher is better

| Training strategy | # of *source* frames at test time | $L1$ error | % Improvement |
|---|---|---|---|
| Training stage **I** | 1 | 0.0632 | 0% |
| Training stage **II** | 1 | 0.0630 | 0.32% |
| Training stage **I** | 3 | 0.0524 | 17.14% |
| Training stage **II** | 3 | 0.0521 | 17.62% |

The results in Table 1 confirm that both training with the curriculum strategy and using additional views at *test* time improve the reconstructed image. Section A.3 in the supplementary material includes qualitative results and shows that using additional *source* frames when testing is especially useful if a face is seen at an extreme pose in the initial *source* frame.

### 5.2    Controlling image generation with a *driving* frame

The motivation of our architecture is to be able to map the expression and pose of a *driving* frame onto a *source* frame *without* any annotations on expression or pose. This section demonstrates that X2Face does indeed achieve this, as a set of *source* frames can be controlled with a driving video and generate realistic results. We compare to two methods: CycleGAN [45] which uses *no* labels and [1] which is designed top down and demonstrates impressive results. Additional

*Source* frames
for X2Face

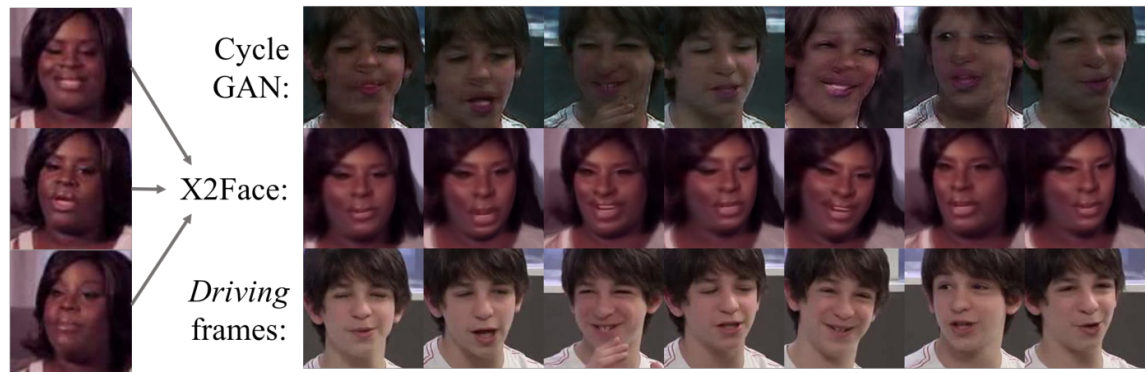

(a)

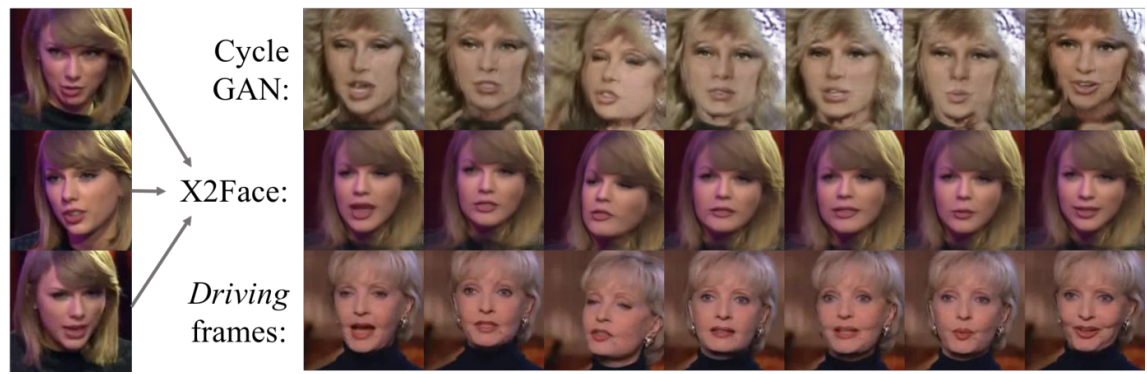

(b)

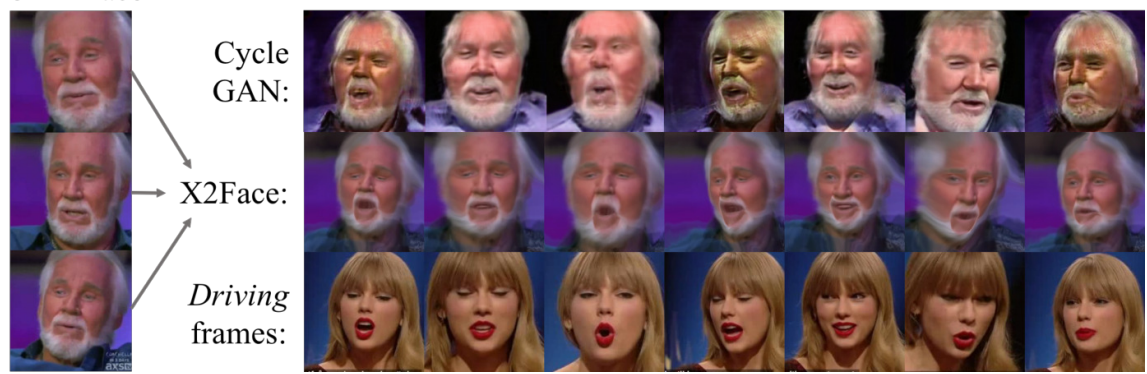

(c)

Fig. 4: Comparison of X2Face's *generated* frames to those of CycleGAN given a driving video sequence. Each example shows from bottom to top: the *driving* frame, our *generated* result and CycleGAN's generated result. To the left, *source* frames for X2Face are shown (at test time CycleGAN does not require *source* frames, as it is has been trained to map between the given *source* and *driving* identities). These examples demonstrate multiple benefits of our method. *First*, X2Face is capable of preserving the face shape of the source identity (top row) whilst driving the pose and expression according to the *driving* frame (bottom row); CycleGAN correctly keeps pose and expression but loses information about face shape and geometry when given too few training images as in example (a) (whereas X2Face requires no training samples for new identities). *Second*, X2Face has temporal consistency. CycleGAN samples from the latent space, so it sometimes samples from different videos resulting in jarring changes between frames (e.g. in example (c)).

qualitative results are given in Fig. 12 in the supplementary material and in the accompanying video[1].

*Comparison to CycleGAN [45].* CycleGAN learns a mapping from a given domain (in this case a given identity A) to another domain (in this case another identity B). To compare to their method for a given pair of identities, we take all images of the given identities (so images may come from different video tracks) to form two sets of images: one set corresponding to identity A and the other to B. We then train their model using these sets. To compare, for a given *driving* frame of identity A, we visualise their *generated* frame from identity B which is compared to that of X2Face.

The results in Fig. 4 illustrate multiple benefits. First, X2Face generalises to unseen pairs of identities at test time given only a *source* and *driving* frame. CycleGAN is trained on pairs of identities, so if there are too few example images, it fails to correctly model the shape and geometry of the *source* face, producing unrealistic results. Additionally, our results have better temporal coherence (i.e. consistent background/hair style/etc. across *generated* frames), as X2Face transforms a given frame whereas CycleGAN samples from a latent space.

*Comparison to Averbuch-Elor et. al. [1].* We compare to [1] in Fig. 5. There are two significant advantages of our formulation over theirs: first, we can handle more significant pose changes in the driving video and *source* frame (Fig. 5b-c). Second, ours has fewer assumptions: (1)[1] assumes that the first frame of the driving video is in a frontal pose with a neutral expression and that the *source* frame also has a neutral expression (Fig. 5d). (2) X2Face can be used when given a single *driving* frame whereas their method requires a video so that the face can be tracked and the tracking used to expand the number of correspondences and to obtain high level details.

While this is not the focus of this paper, our method can be augmented with the ideas from these methods. For example, as inspired by [1], we can perform simple post-processing to add higher level details (Fig. 5a, X2Face+p.p.) by transferring hidden regions using Poisson editing [28].

### 5.3   Controlling the image generation with pose

Before reporting results on controlling the *driving vector* using pose, we validate our claim that the *driving vector* does indeed learn about pose. To do this, we evaluate how accurately we can predict the three head pose angles – yaw, pitch and roll – given the 128D *driving vector*.

*Pose predictor.* To train the pose predictor which also serves as $f_{v \rightarrow p}$ (Section 4.1), the $25,993$ images in the AFLW dataset [19] are split into train/val set, leaving out the $1,000$ test images from [22] as test set. The results on the test set are reported in Table 2 confirming that the *driving vector* learns about head

---

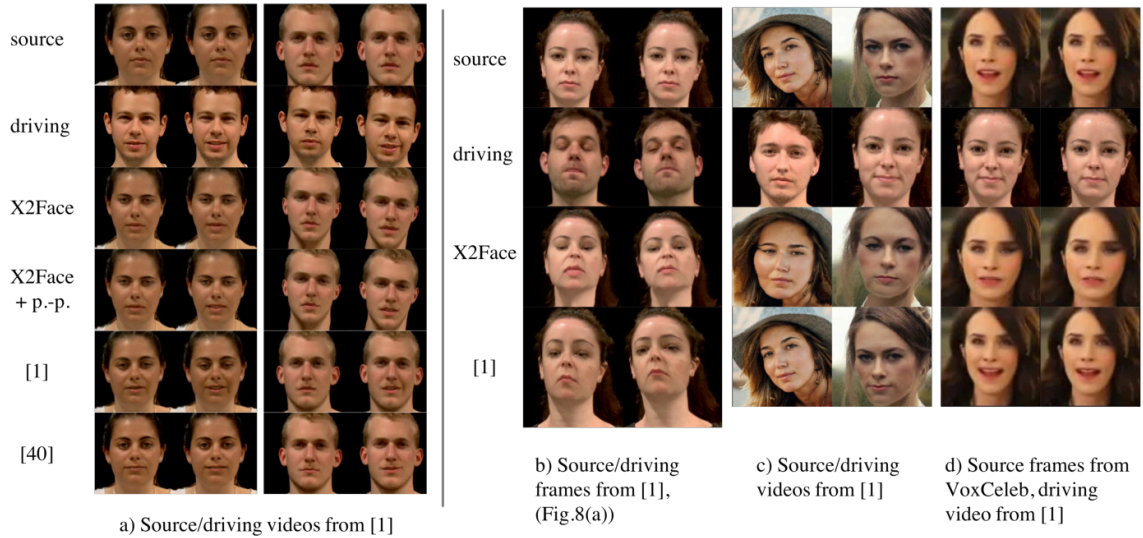

a) Source/driving videos from [1]

b) Source/driving frames from [1], (Fig.8(a))

c) Source/driving videos from [1]

d) Source frames from VoxCeleb, driving video from [1]

Fig. 5: Comparison of X2Face to supervised methods. In comparison to [1]: X2Face matches (b) pitch, and (c) roll and yaw; and X2Face can handle non-neutral expressions in the *source* frame (d). As with other methods, post-processing (X2Face + p.-p.) can be applied to add higher level details (a).

pose without having been trained on pose labels, as the results are comparable to those of a network directly trained for this task.

We then use $f_{v \to p}$ to train $f_{p \to v}$ (Section 4.1) and present generated frames for different, unseen test identities using the learnt mappings in Fig. 6. The *source* frame corresponds to $p_{source}$ in Section 4.1 while $p_{driving}$ is used to vary one head pose angle while keeping the others fixed.

Table 2: MAE in degrees using the *driving vector* for head pose regression (lower is better). Note that the linear pose predictor from the *driving vector* performs only slightly worse than a supervised method [22], which has been trained for this task

| Method | Roll | Pitch | Yaw | MAE |
|---|---|---|---|---|
| X2Face | 5.85 | 7.59 | 14.62 | 9.36 |
| KEPLER [22] (supervised) | 8.75 | 5.85 | 6.45 | 7.02 |

## 5.4   Controlling the image generation with audio input

This section presents qualitative results for using audio data from videos in the VoxCeleb dataset to drive the *source* frames. The VoxCeleb dataset consists of videos of interviews, suggesting that the audio should be especially correlated with the movements of the mouth. [9]'s model, trained on the BBC-Oxford 'Lip

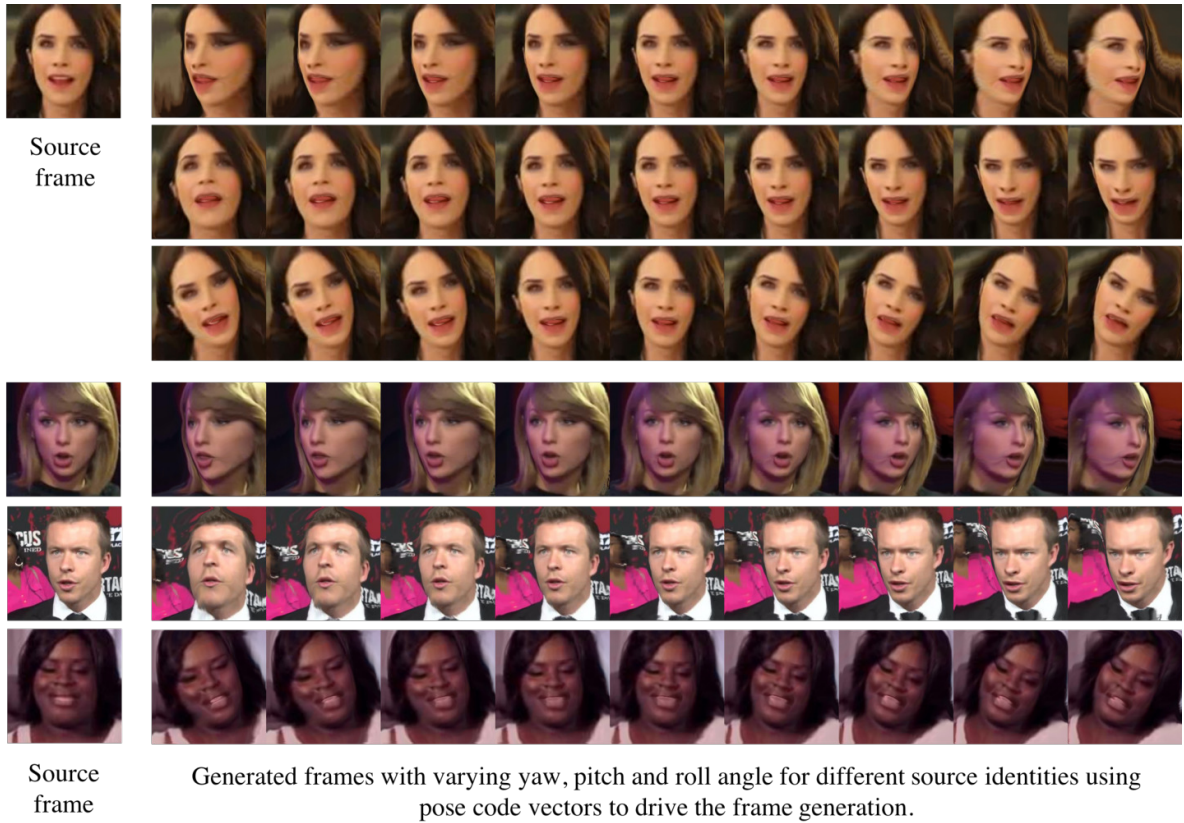

Source
frame

Generated frames with varying yaw, pitch and roll angle for different source identities using
pose code vectors to drive the frame generation.

Fig. 6: Controlling image generation with pose code vectors. Results are shown for a single *source* frame which is controlled using each of the three head pose angles for the same identity (top three rows) and for different identities (bottom three rows). For further results and a video animation, we refer to Fig. 15 in the supplementary material and to the accompanying video. Whilst some artefacts are visible, the method allows the head pose angles to be controlled separately.

Reading in the Wild' dataset (LRW), is used to extract audio features. We use the 256D vector activations of the last fully connected layer of the audio stream (FC7) for a 0.2s audio signal centred on the *driving* frame (the frame occurs half way through the 0.2s audio signal).

A potential source of error is the domain gap between the LRW dataset and VoxCeleb, as [9]'s model is not fine-tuned on the VoxCeleb dataset which contains much more background noise than the LRW dataset. Thus, their model has not necessarily learnt to become indifferent to this noise. However, our model is relatively robust to this problem; we observe that the mouth movements in the *generated* frames are reasonably close to what we would expect from the sounds of the corresponding audio, as demonstrated in Fig. 7. This is true even if the person in the video is not speaking and instead the audio is coming from an interviewer. However, there is some jitter in the generation.

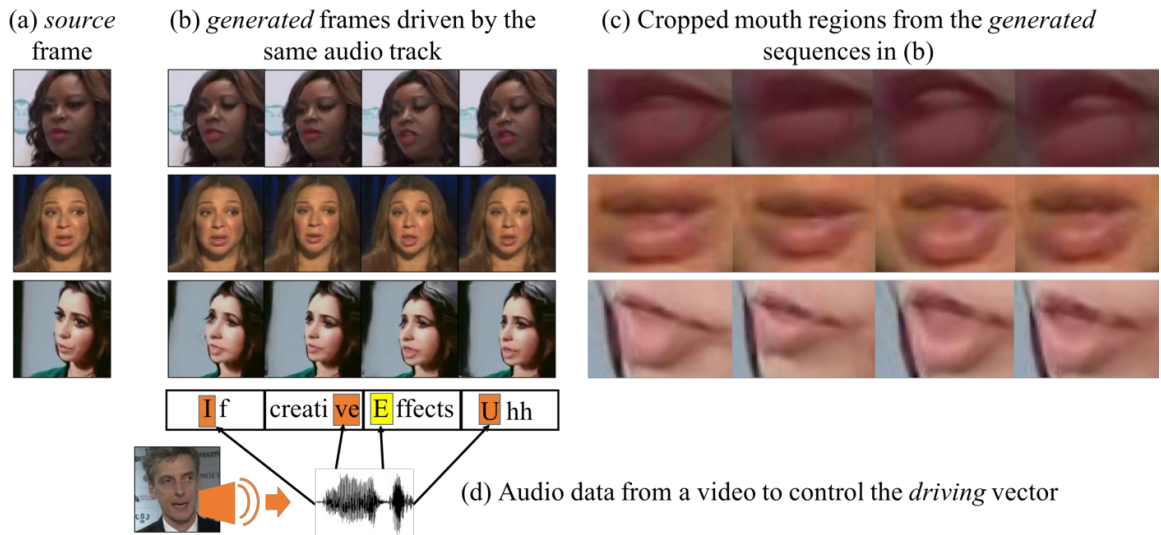

Fig. 7: Controlling image generation with audio information. We show how the same sounds affect various *source* frames; if our model is working well then the *generated* mouths should behave similarly. (a) shows the *source* frames. (b) shows the *generated* frames for a given audio sound which is visualised in (d) by the coloured portion of the word being spoken. As most of the change is expected to be in the mouth region, the cropped mouth regions are additionally visualised in (c). The audio comes from a native British speaker. As can be seen, in all generated frames, the mouths are more closed at the "ve" and "I" and more open at the "E" and "U". Another interesting point is that for the "Effects" frame, the audio is actually coming from an interviewer, so while the frame corresponding to the audio has a closed mouth, the *generated* results still open the mouth.

## 6    Using the embedded face for video editing

We consider how the *embedded* face can be used for video editing. This idea is inspired by the concept of an unwrapped mosaic [31]. We expect the *embedded* face to be pose and expression invariant, as can be seen qualitatively across the example *embedded* faces shown in the paper. Therefore, the *embedded* face can be considered as a UV texture map of the face and drawn on directly.

This task is executed as follows. A *source* frame (or set of *source* frames) is extracted and input to the *embedding network* to obtain the *embedded* face. The *embedded* face can then be drawn on using an image or other interactive tool. A video is reconstructed using the modified *embedded* face which is driven by a set of *driving* frames. Because the *embedded* face is stable across different identities, a given edit can be applied across different identities. Example edits are shown in Fig. 8 and in Fig. 13 in the supplementary material.

## 7    Conclusion

We have presented a self-supervised framework X2Face for driving face generation using another face. This framework makes no assumptions about the

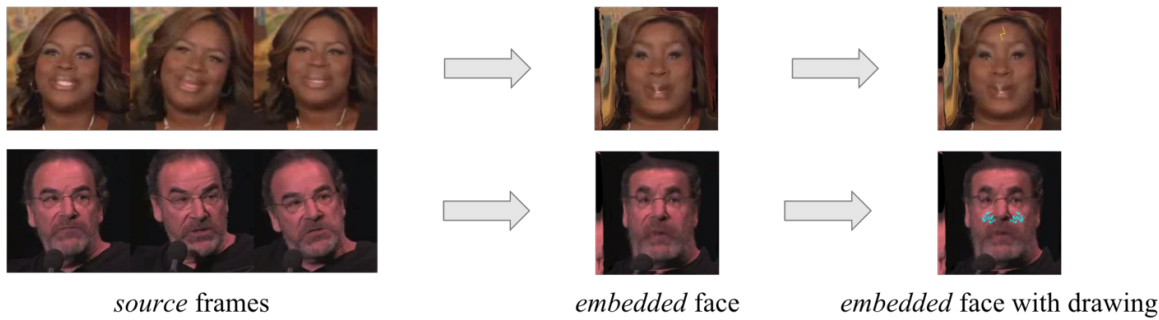

(a) *Source* frames are input to extract the *embedded* face which is drawn on. The modified *embedded* face is used to *generate* the frames below.

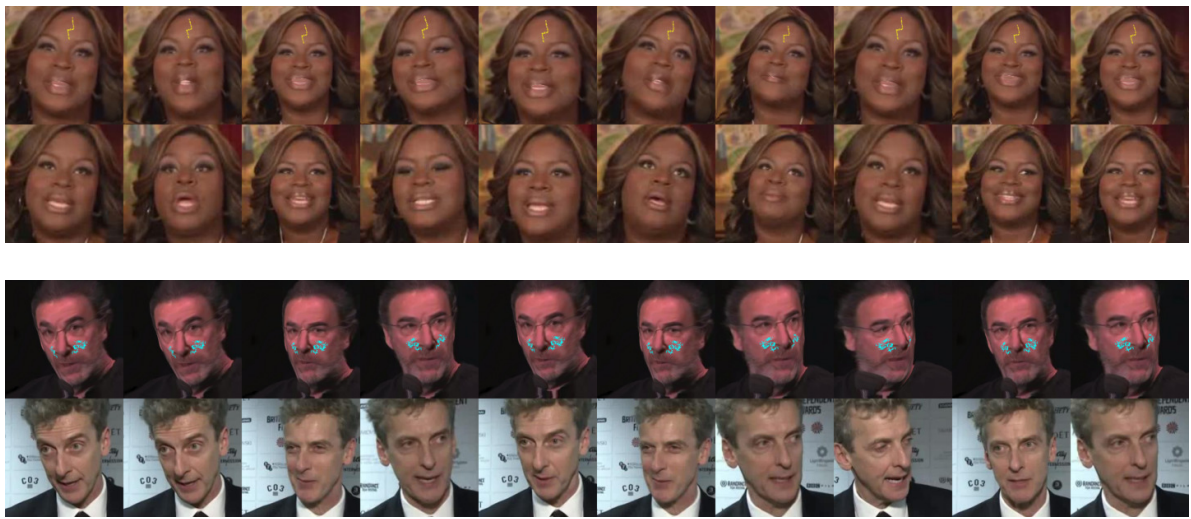

(b) An example sequence of *generated* frames (top row) from the modified *embedded* face controlled using a sequence of *driving* frames (bottom row).

Fig. 8: Example results of the video editing application. (a) For given *source* frames, the *embedded* face is extracted and modified. (b) The modified *embedded* face is used for a sequence of *driving* frames (bottom) and the result is shown (top). Note how for the second example, the blue tattoo disappears behind the nose when the person is seen in profile and how, as above, the modified *embedded* face can be driven using the same or another identity's pose and expression. Best seen in colour. Zoom in for details. Additional examples using the blue tattoo and Harry Potter scar are given in the accompanying video and in Fig. 13 in the supplementary material.

pose, expression, or identity of the input images, so it is more robust to unconstrained settings (e.g. an unseen identity). The framework can also be used with minimal alteration *post* training to drive a face using audio or head pose information. Finally, the trained model can be used as a video editing tool. Our model has achieved all this without requiring annotations for head pose/facial landmarks/depth data. Instead, it is trained self-supervised on a large collection of videos and learns itself to model the different factors of variation.

While our method is robust, versatile, and allows for generation to be conditioned on other modalities, the generation quality is not as high as approaches specifically designed for transforming faces (e.g. [1,17,40]). This opens an interesting avenue of research: how can the approach be modified such that the versatility, robustness, and self-supervision aspects are retained but with the generation quality of these methods that are specifically designed for faces. Finally, as no assumptions have been made that the videos are of faces, it is interesting to consider applying our approach to other domains.

**Acknowledgements** The authors are grateful to Hadar Averbuch-Elor for helpfully running their model on our data and to Vicky Kalogeiton for suggestions/comments. This work was funded by an EPSRC studentship and EPSRC Programme Grant Seebibyte EP/M013774/1.

## Supplementary Material

We provide additional details about X2Face's architecture and the training and testing setup in Section A, and more qualitative results in Section B.

## A    Additional details on architectures and training.

Section A.3 provides qualitative results on using additional views at *test* time and the curriculum strategy. Section A.1 and Section A.2 give additional details about X2Face's two subnetworks: the *embedding network* and the *driving network*.

### A.1    Embedding network

The *embedding network* is based on the pix2pix [15] version of the U-Net architecture [32], except that the last layer is changed to a 2-channel layer. The resulting $2 \times 256 \times 256$ gives the sampler which determines how to sample from the *source* frame to obtain the *embedded* face. The exact architecture is given in Fig. 9.

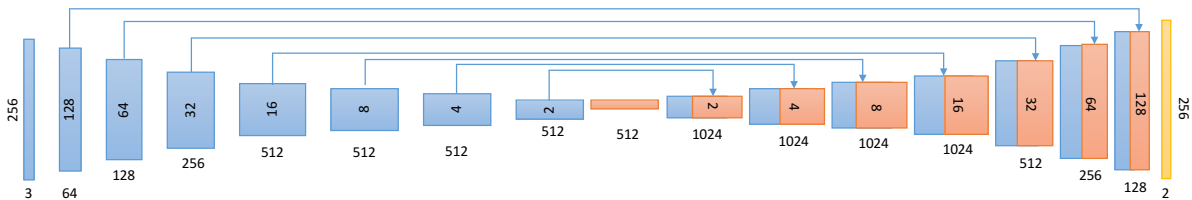

Fig. 9: *Embedding Network*: In the encoder, each convolution is followed by a leaky ReLU with factor 0.2 and then a batch-norm layer (except for the first which has no batch-norm). The convolutional filter sizes are $4 \times 4$ and the stride/padding is 2/1. In this decoder, the layers from the encoder are concatenated using skip connections, doubling the channel size. In the decoder, the sequence of executions following each convolutional layer is: ReLU, bilinear upsampling, batch-norm, concatenation (with the corresponding encoder layer). The last layer (yellow) does not have a concatenation. The final $2 \times 256 \times 256$ result is passed through a tanh layer to give X2Face's prediction for how to sample from the *source* frames. The convolutional filter sizes in the decoder are $3 \times 3$ and the stride/padding is 1/1.

### A.2    Driving network

The *driving network* is an encoder-decoder network that uses the same layer sizes as the *embedding network* but does not have any skip connections. The *driving vector* corresponds to the vector resulting from the encoder portion of the network, which is a 128D vector. The architecture is shown in Fig. 10.

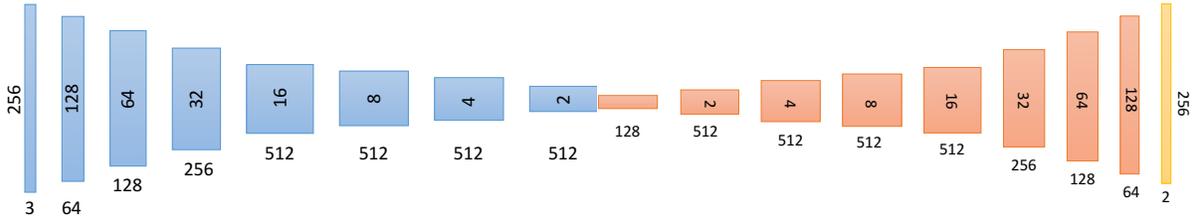

Fig. 10: *Driving Network*: The encoder is similar to the encoder of the *embedding network*; each convolution is followed by a leaky ReLU (factor 0.2) and then a batch-norm layer (except for the first which has no batch-norm); the convolutional filter sizes are $4 \times 4$ and the stride/padding is $2/1$. In this decoder, there are no skip connections. The *driving vector* has channel size 128. In the decoder, the sequence of executions following each convolutional layer is: ReLU, bilinear upsampling, batch-norm. The final $2 \times 256 \times 256$ result is passed through a tanh layer to give X2Face's prediction for how to sample from the *embedded* face. The convolutional filter sizes in the decoder are $3 \times 3$ and the stride/padding is $1/1$.

### A.3   Qualitative results for different training/testing setups

We consider the following 4 different training/testing settings:

1. **Training stage I, Single source (S):** Training with photometric $L1$ loss only and testing with a single *source* frame.
2. **Training stage I, Multi-source (M):** Training with photometric $L1$ loss only and testing with multiple *source* frames.
3. **Training stage II, S:** Training with photometric $L1$ loss and identity losses as described in Section 3.2 in the paper and testing with a single *source* frame.
4. **Training stage II, M:** Training with photometric $L1$ loss and identity losses as described in Section 3.2 in the paper and testing with multiple *source* frames.

As can be seen in Fig. 11, testing with multiple *source* frames makes our method more robust, as the *generated* frame is not so unstable with respect to the choice of the *source* frame. This improves the quality and sharpness of the *generated* frames.

Although the results get better quantitatively when adding the identity loss functions (as discussed in the paper), we observe that our method produces visually very similar and convincing results when trained with the photometric $L1$ loss only. The use of the identity content losses is the only point where we make use of some form of labelled supervision, as we use a pre-trained network that was trained for identity classification. Using only the photometric loss results in a *completely self-supervised* framework.

## B   Additional qualitative results

We present more qualitative results comparing X2Face to CycleGAN [45] in Fig. 12 and show additional results for controlling the head pose of a *source*

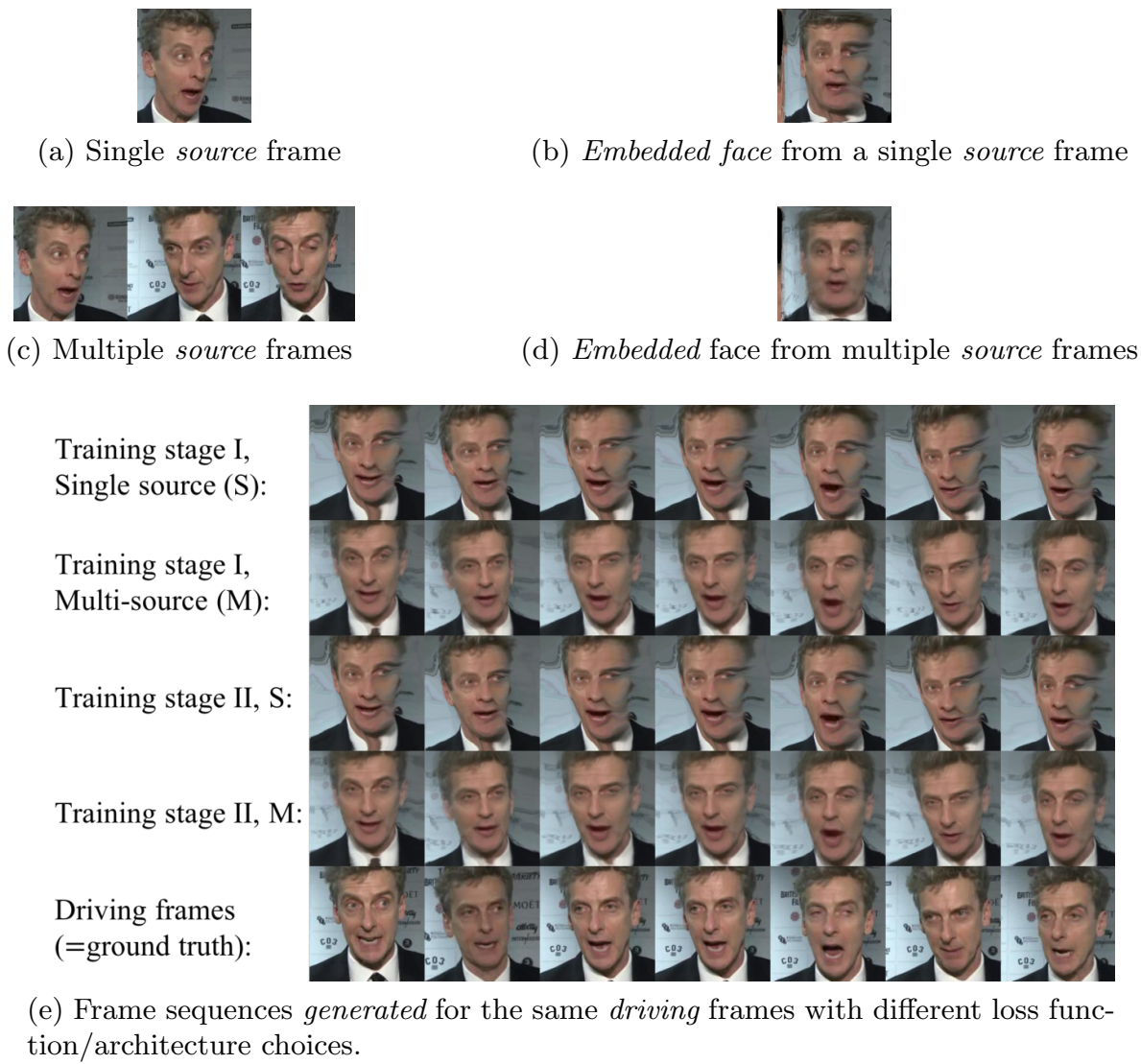

(a) Single *source* frame

(b) *Embedded face* from a single *source* frame

(c) Multiple *source* frames

(d) *Embedded* face from multiple *source* frames

Training stage I,
Single source (S):

Training stage I,
Multi-source (M):

Training stage II, S:

Training stage II, M:

Driving frames
(=ground truth):

(e) Frame sequences *generated* for the same *driving* frames with different loss function/architecture choices.

Fig. 11: A comparison of different training/testing settings. (a) The *source* frame used when testing using a single *source* frame. (b) The *embedded* face for this single *source* frame. (c) The *source* frames used when testing with multiple *source* frames. (d) The *embedded* face from these multiple *source* frames. (e) *Generated* frames for different training stages and testing strategies. As the *driving* frame is from the same video, the *driving* frame corresponds to the ground truth and is shown in the bottom row. When using a single *source* frame for testing, the model is sensitive to the quality/pose of the *source* frame (see rows for *Training stage I, Single source (S)* and *Training stage II, S*). Using additional *source* frames and averaging over the resulting *embedded* face produces a better *embedded* face representation and higher quality results (see rows for *Training stage I, Multi-source (M)* and *Training stage II, M*).

frame using a pose vector in Fig. 15. Finally, additional qualitative results for using X2Face for video editing are given in Fig. 13 and Fig. 14.

For further demonstrations of X2Face in action, we refer to the accompanying video which can be found at http://www.robots.ox.ac.uk/~vgg/research/unsup_learn_watch_faces/x2face.html.

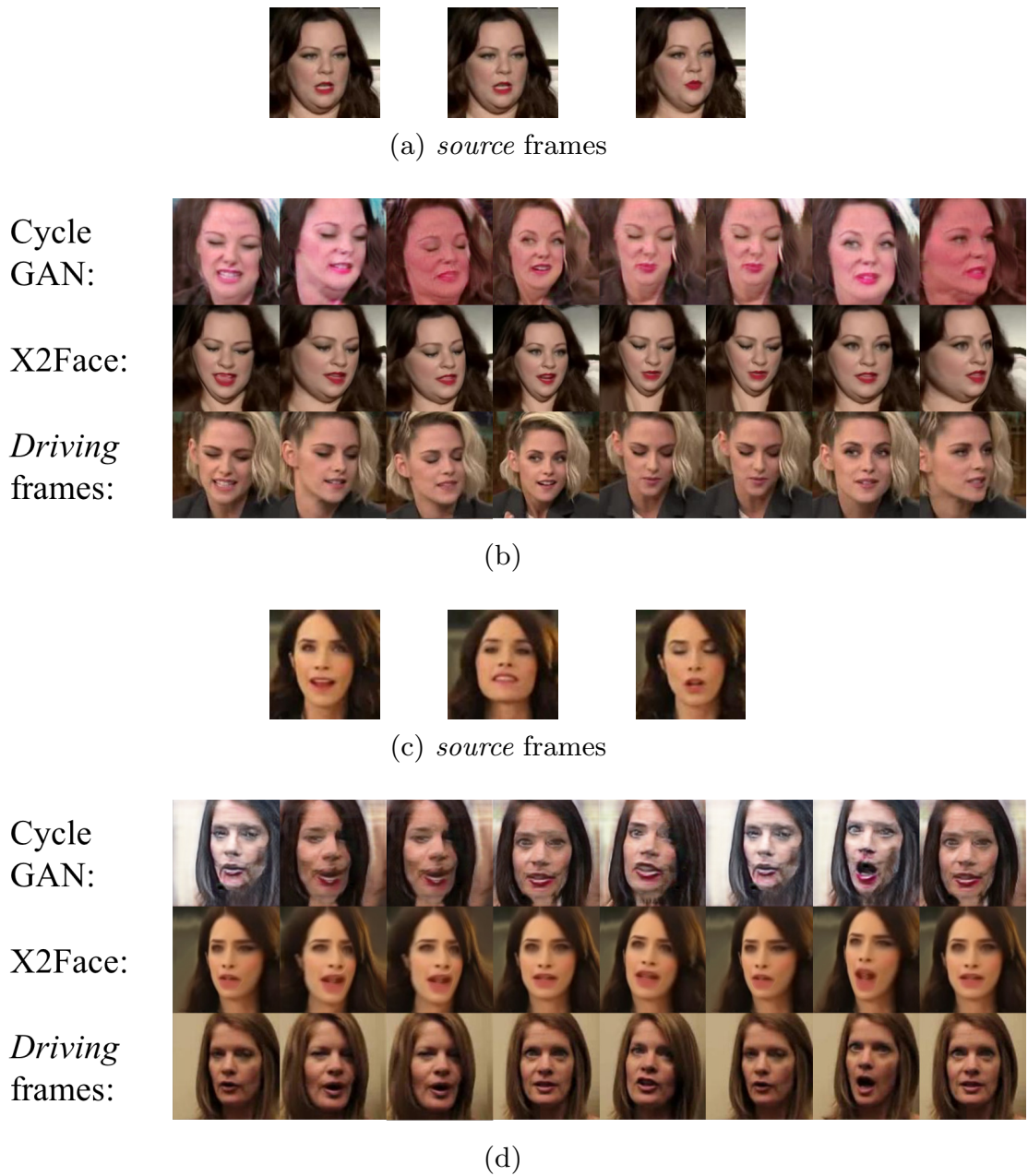

Fig. 12: More qualitative results comparing X2Face to CycleGAN. Note how X2Face preserves the hair/background setting from the *source* frames, resulting in temporal coherence across the *generated* frames.

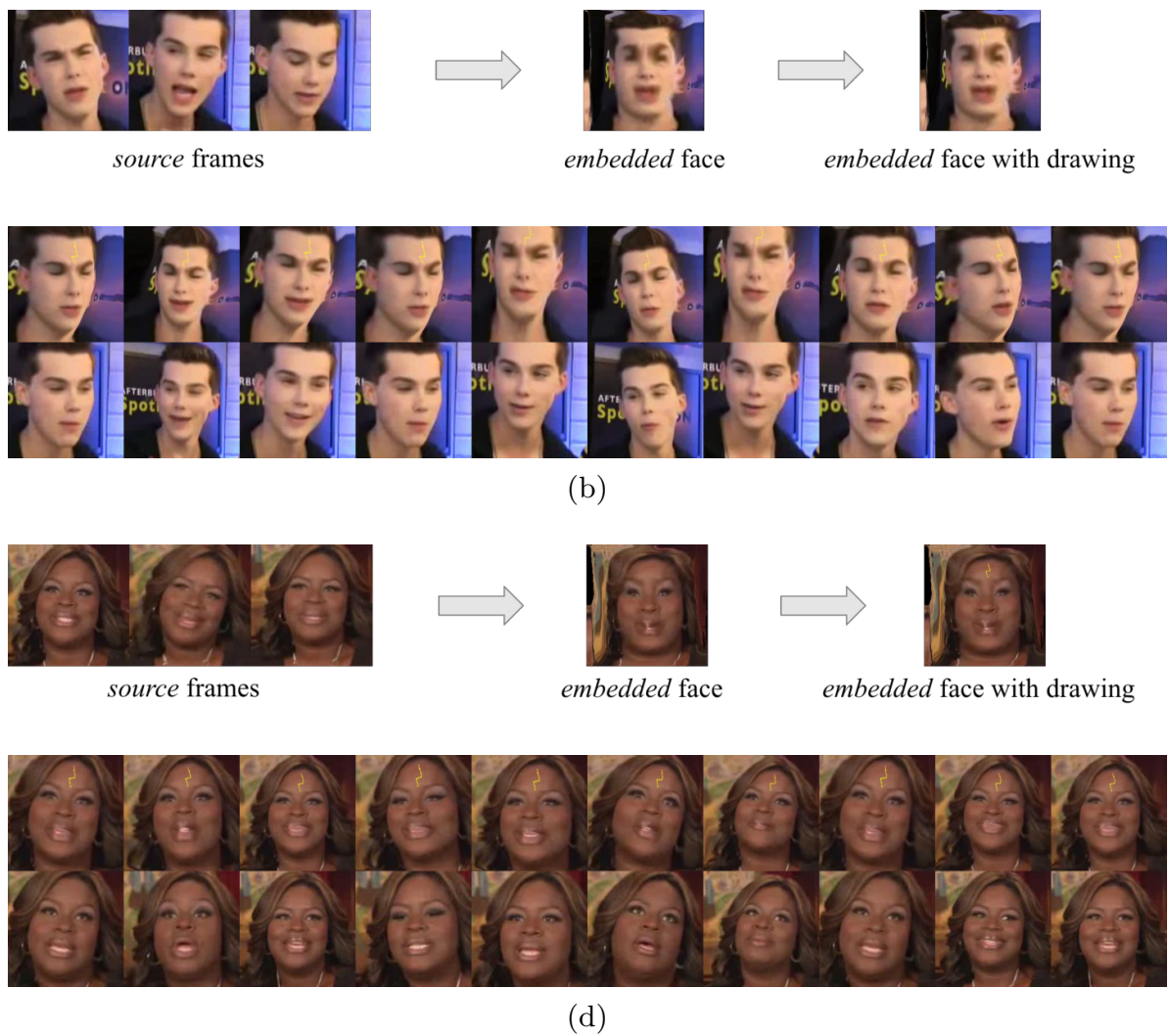

Fig. 13: Additional results of the video editing application using the Harry Potter scar. (a) For the given *source* frames, the *embedded* face is extracted and modified. (b) The modified *embedded* face is controlled using a sequence of *driving* frames (bottom) and the result is shown (top). Best viewed in colour. Zoom in for details.

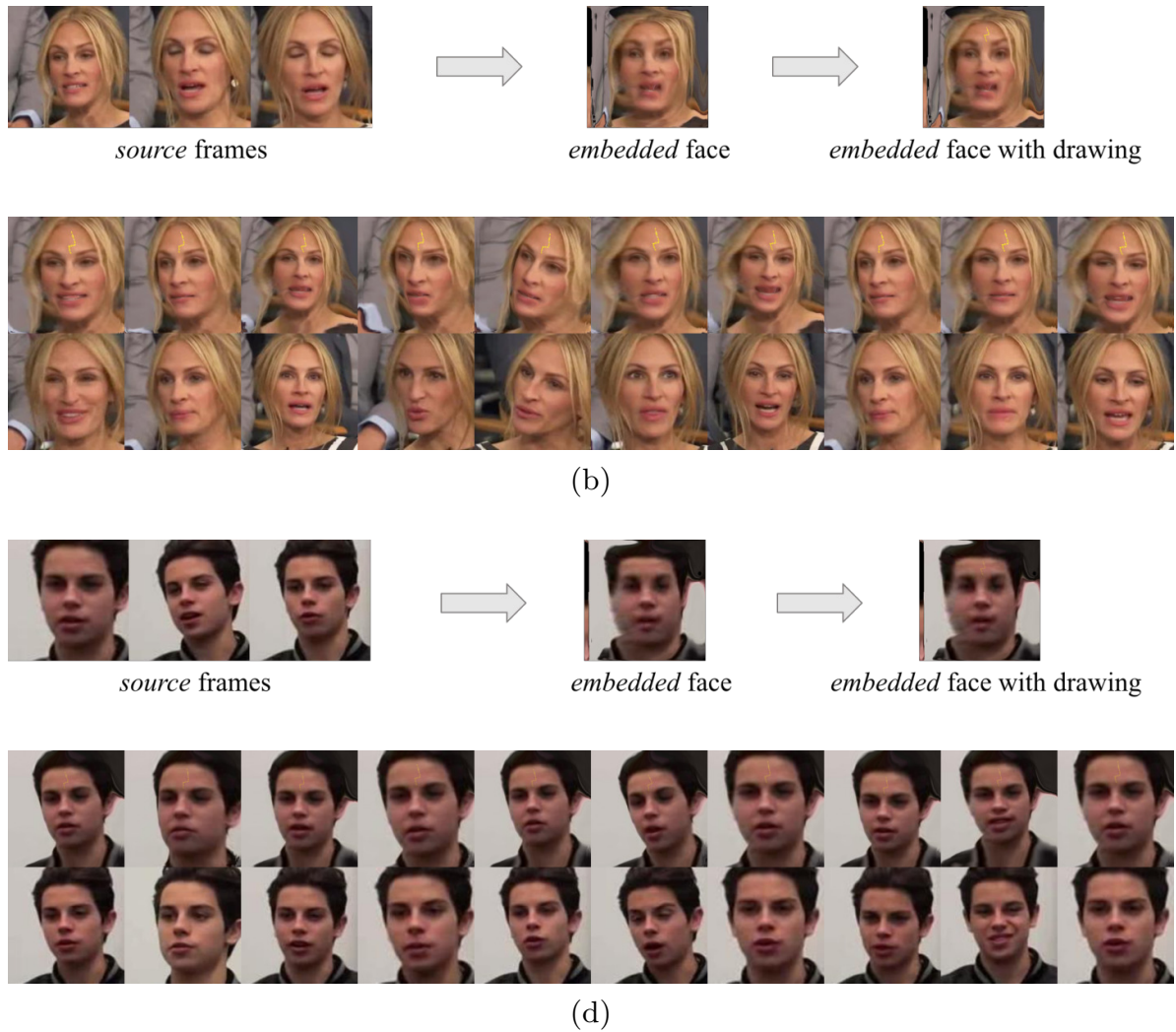

(b)

(d)

Fig. 14: Additional results of the video editing application using the Harry Potter scar. (a) For the given *source* frames, the *embedded* face is extracted and modified. (b) The modified *embedded* face is controlled using a sequence of *driving* frames (bottom) and the result is shown (top). Best viewed in colour. Zoom in for details.

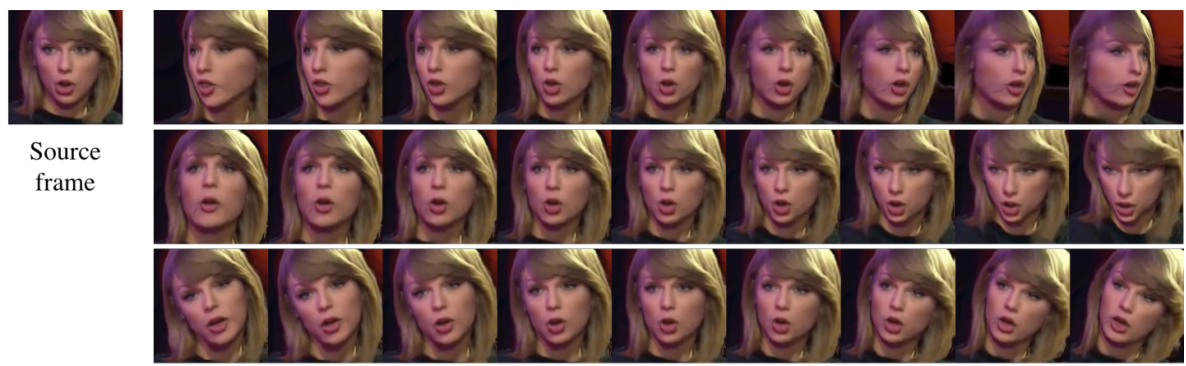

Generated frames with varying yaw, pitch and roll angle (top row, middle row and bottom row)
using pose code vectors to drive the frame generation.

Fig. 15: Controlling the three head pose angles. This example can also be found
in the accompanying video.